\documentclass[letterpaper, 10 pt, conference]{ieeeconf}  

\IEEEoverridecommandlockouts                              

\usepackage{hyperref}
\usepackage{graphicx}
\usepackage{algorithm}
\usepackage{algpseudocode}  
\usepackage{soul} 
\soulregister{\hl}{1} 
\usepackage{float}          
\usepackage{amsmath}
\usepackage{amsfonts} 
\usepackage{multirow} 
\usepackage{soul}
\usepackage{capt-of,etoolbox}
\usepackage{makecell}
\usepackage{float}
\usepackage{color,soul}
\let\labelindent\relax
\usepackage{enumitem}

\title{\LARGE \bf
Seeing Fast and Slow: \\
Bimodal 3D Scene Graphs for Open-set Tasks
}

\author{Marcel Bartholomeus Prasetyo$^{\dagger}$, Shrutika Vishal Thengane$^{\dagger}$, 
A Manicka Praveen$^{\dagger}$, Yi Loo, Malika Meghjani \\
{\small Singapore University of Technology and Design (SUTD), Singapore}}

\begin{document}

\twocolumn[{%
\renewcommand\twocolumn[1][]{#1}%
\maketitle

\begin{center}
    \centering
    \includegraphics[width=\linewidth]{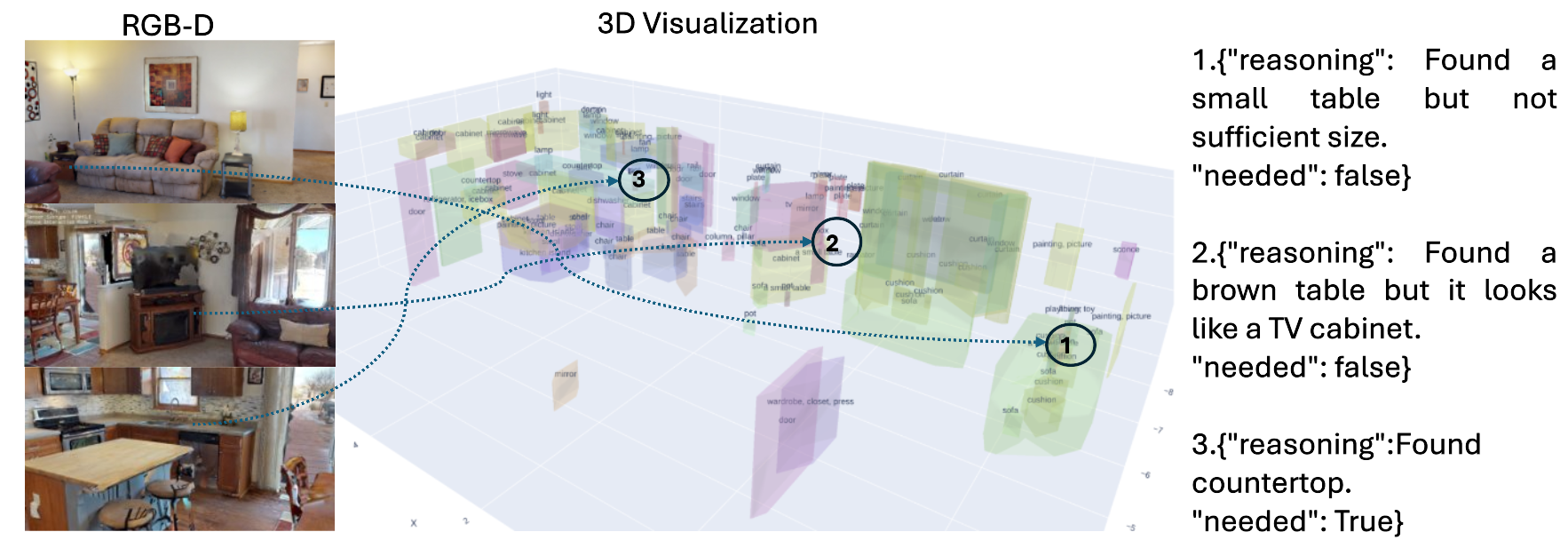} 
    \vspace{-0.03\textheight}
    \captionof{figure}{Bimodal 3D scene graph generated for the complex object retrieval task, ``I just bought vegetables and milk, what should I do?'' The default ``fast'' mode identifies closed-set objects relevant for the task. Mode switching then triggers ``slow'' mode to further investigate candidate objects via VLM description and LLM verification, confirming their relevance to the task. The relevant objects are annotated with their respective reasoning.}
    \label{fig:open-example}
\end{center}
}]

\begingroup
\renewcommand\thefootnote{}%
\footnotetext{$^{\dagger}$ denotes equal contribution in no particular order}%
\footnotetext{This work was supported by the TL@SUTD Seed Research Programme.}%
\endgroup

\thispagestyle{empty}
\pagestyle{empty}

\begin{abstract}

Open-set task execution can significantly benefit in real-time reasoning and planning by seamlessly switching between coarse and fine scene representations, given the context and the evolving information. For example, it is often sufficient to start with a coarse scene representation initially and only employ a finer, more granular scene representation when the robot encounters regions which are likely to contain the task relevant objects. Hence, in this work, we propose BiMoSG, a bimodal 3D scene graph generation approach for open-set task execution. 
BiMoSG employs a "fast" mode by default to efficiently generate a coarse 3D scene graph and can switch to a "slow" mode for generating a finer open vocabulary 3D scene graph of task relevant objects, where the mode switching is based on reasoning of task related efficiencies. 
We demonstrate that our proposed 3D scene graph approach can be deployed on mobile robots with the capability of generating real-time semantically and geometrically aligned scene graphs for performing simple as well as complex open-set tasks in unknown environments.

\end{abstract}

\section{Introduction}

Open-set tasks in unknown environments require reasoning and scene understanding in real-time. Thus, the visual representations of the environment would also need to be constructed and updated in real-time by the robot.
Recently, open vocabulary 3D scene graphs \cite{gu2024conceptgraphs,maggio2024clio,Werby-RSS-24, koch2024open3dsg} have emerged as an effective class of methods to represent 3D scenes for robot navigation and related tasks. These methods utilize large visual foundation models such as SAM \cite{kirillov2023segment}, CLIP \cite{radford2021learning} and LLaVa \cite{liu2023llava} to achieve both robust instance segmentation as well as open-set classification of objects. However, generating 3D scene graph in real-time on mobile robots remains challenging due to the substantial computational demands of visual foundation models paired with the limited computational resources on the edge. Additionally, the high energy cost of these models may also prevent robots from operating for an extended time-span, which is often required for open-set task execution.

We posit that for open-set task execution, it is not necessary for the robot to generate high-granularity, open-vocabulary 3D scene graphs for the majority of the duration of the task. 
For example, as shown in Fig. \ref{fig:open-example}, consider a robot that is tasked with taking care of recently bought groceries. 
As the robot navigates through the room, it only requires coarse semantic information regarding objects to determine their general functionality (e.g., small tables and TV cabinets are not appropriate for storing or organizing the groceries). Only when the robot encounters an object with a high likelihood of relevance to the task of interest (e.g., countertop for storage) does it require switching to a finer scene representation to precisely execute the desired task. This is also partly inspired by how humans also possess a fast, coarse System 1 thinking, and a slow, deliberate System 2 thinking \cite{kahneman2011thinking} but applied to a robot's perception system.

Guided by this insight, we propose BiMoSG, a fast 3D scene graph framework with two complementary modes. The default \textit{Fast Mode} constructs a coarse scene representation using a closed-set vocabulary, a lightweight prismatic representation of 3D objects, and vicinity-based spatial hierarchies. This provides sufficient geometric and semantic context for efficient scene understanding and task execution without relying on large visual foundation models.
For tasks requiring open-vocabulary retrieval or visual reasoning, BiMoSG can selectively invoke a \textit{Mode Switching}, which builds upon the fast representation and performs targeted open-set investigation. The two modes therefore complement one another: Fast Mode provides efficient continuous scene mapping, while Switching Mode allocates additional visual and language reasoning only when required by the task.

The main contributions of our work are as follows:
(a) a novel volumetric representation of 3D objects which helps in generating fast 3D scene graph representations,
(b) a novel bi-modal switching approach for generating task specific resolution of the scene graph based on reasoning,
(c) a real-time on-the-edge framework (BiMoSG) for generating scene graphs in unknown environments.

\begin{figure*}[ht]
    \centering
    \includegraphics[width=.8\linewidth]{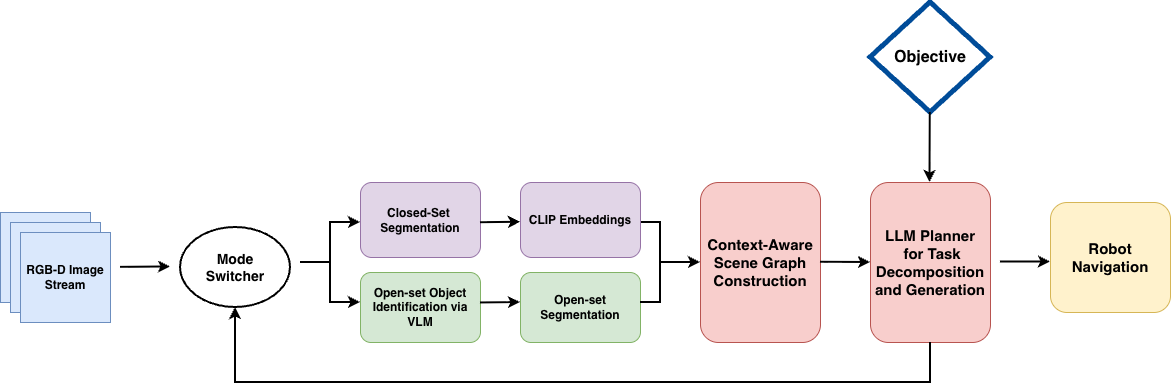} 
    \caption{Overview of proposed approach. Our system uses the closed segmentation using Mask2Former by default to generate a coarse scene graph. If the LLM Planners detects objects that matches one of the triggered object labels, it can trigger the "slow" open-set object identification module using VLM to perform open-set segmentation. The identified open-set objects are then appended to the existing scene graph.} 
    \label{fig:proposed_method}
\end{figure*}


\vspace{-0.005\textheight}

\section{Related Work}

Recent work in 3D scene visual understanding leverages large vision-language models (LVLMs) to construct 3D scene representations \cite{gorlo2025moment, jatavallabhula2023conceptfusion, peng2023openscene}. These methods typically ground 2D representations generated by LVLMs into the 3D scene. A key advantage of LVLMs is their ability to provide open-vocabulary representations of 3D scenes. However, the computational requirements of LVLMs can make them impractical for real-time deployment on mobile robots. Another class of 3D scene graph approaches combines classical robotics approaches with deep learning and foundation models for real-time deployment on mobile robots. Our work focuses on this class of scene graph approaches, with particular emphasis on open-set semantics. The components of this class of scene graphs include geometrical representations, segmentation, semantics and graph hierarchy. Thus, we compare our approach with the state-of-the-art approaches in terms of these components.

\subsubsection{\textbf{Geometrical Representation}}
Concept Graphs \cite{gu2024conceptgraphs}, HOV-SG \cite{Werby-RSS-24}, and BBQ \cite{linok2025beyond} make use of point clouds, clustered per object and sometimes voxelized. Hydra \cite{hughes2022hydra} and Kimera \cite{rosinol2020kimera} make use of 3D mesh representations. We aim to introduce a prismatic volumetric geometry representation for our approach to obtain a compact, computationally efficient representation of an object's 3D spatial extent that is easier to use for geometric and spatial reasoning.

\subsubsection{\textbf{Segmentation}}

Most open-set scene graph approaches employ SAM or faster alternatives such as FastSAM \cite{zhao2023fast} and MobileSAM \cite{zhang2023faster}. We instead adopt a two-stage segmentation strategy, using Mask2Former \cite{cheng2022maskedattentionmasktransformeruniversal} for closed-set segmentation and SAM3 \cite{carion2026sam} for open-set segmentation. Mask2Former provides reliable instance masks across a broad set of common object categories, while SAM3 extends segmentation to objects outside the predefined category set.

\subsubsection{\textbf{Open-set Semantics}}

HOV-SG \cite{Werby-RSS-24} and BBQ \cite{linok2025beyond} leverage VLMs, while ConceptGraphs \cite{gu2024conceptgraphs}, Clio \cite{maggio2024clio}, and ASHiTA \cite{chang2025ashita} use CLIP for open-set semantic representation. We combine both paradigms in an adaptive pipeline: CLIP provides lightweight task-conditioned semantic matching, while SmolVLM2 \cite{marafioti2025smolvlm} is invoked for more complex cases requiring semantic verification or reasoning. This enables open-set understanding while avoiding VLM inference when lightweight matching is sufficient.

\subsubsection{\textbf{Graph Hierarchy}}

It is desirable for 3D scene graphs to encode hierarchical structure, particularly in terms of spatial organization and semantic granularity. HOV-SG \cite{Werby-RSS-24} organizes the scene graph hierarchically by rooms, while Hydra introduces multiple spatial layers spanning objects, places, rooms, and buildings. We instead introduce a local, decentralized spatial hierarchy based on vicinities, together with a room level proximal hierarchy, enabling objects to be organized according to their local spatial context. Clio \cite{maggio2024clio} and ASHiTA \cite{chang2025ashita} further introduce task driven hierarchies that adapt the computation, task structure, and scene graph granularity to the task. Similarly, we incorporate multiple semantic granularity levels, combining coarse and fine representations across closed-set and open-set semantics, allowing the scene graph to support different levels of task specific semantic understanding.

\section{Methodology}

BiMoSG is a framework for efficient environment exploration and semantic 3D scene graph construction, as illustrated in Fig.~\ref{fig:proposed_method}. It combines closed-set and open-set segmentation to support both known and unseen object categories. The framework consists of three components: (a) \textbf{Scene Graph Representation}, which performs object segmentation, volumetric representation, duplicate merging, and vicinity-based relation extraction to construct a closed-set 3D scene graph; (b) \textbf{Hybrid Representation}, which augments this scene graph with lightweight, always-on open-set matching, extending fast mode to open-set retrieval tasks; and (c) \textbf{Mode Switching}, a slower, LLM- and VLM-driven planner that escalates beyond the hybrid representation when a task requires further reasoning, such as verifying an ambiguous match or reasoning over a VLM-described scene before confirming an object. Together, these components allow BiMoSG to resolve simple retrieval tasks efficiently within fast mode, while reserving slower, more expressive reasoning for complex tasks that require it.

\vspace{-0.005\textheight}

\subsection{\textbf{Scene Graph Representation}}

\subsubsection{\textbf{Object Segmentation}}
Proposed method employs a dual mode object segmentation by combining the open-set and closed-set segmentation models. In the open-set segmentation, it uses the VLM model to generate the caption (labels) for the newly generated scene objects whose caption is not defined in a closed-set, and generate a mask for the new objects using a language-based segmentation model. For the closed-set segmentation, we are using a mask2former \cite{cheng2022maskedattentionmasktransformeruniversal}, in particular one trained on the ADE20K-150 SceneParse dataset having 150 distinct "things" and "stuff" classes.

\subsubsection{\textbf{Prismatic 3D object Representation}}
For the 3D object representation, we utilise RGB-D images, robot trajectory and camera pose to extract spatial information about objects in the scene. Specifically, we first generate object masks using closed-set segmentation model \cite{cheng2022maskedattentionmasktransformeruniversal}, project the points received onto the world frame and then obtain the 3D object representation of each object before placing them in the scene graph.
\begin{figure}[ht]
    \centering
    \includegraphics[width=\linewidth]{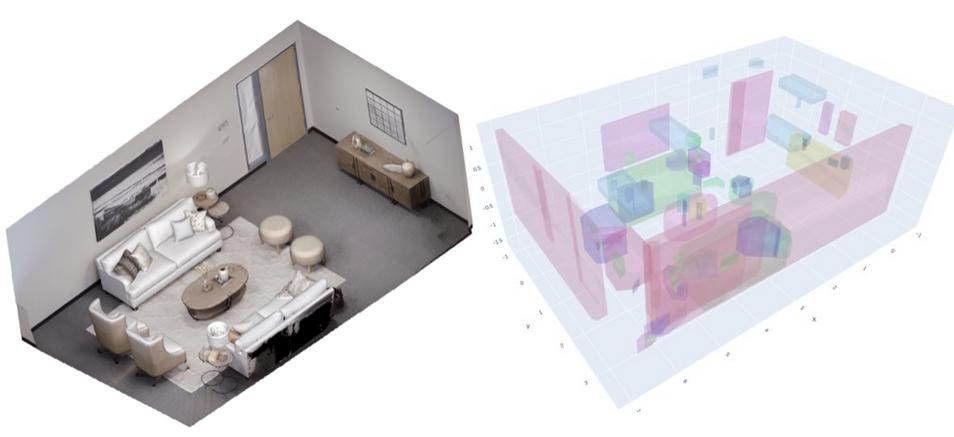} 
    \caption{Prismatic volumetric object representation on replica data(room0).} 
    \label{fig:volumetricrepre}
\end{figure}

RGB-D approaches usually take the segmentation masks of each detected object in the image and project a dense point cloud of the object from image to world coordinates. This ensures high visual fidelity object representation in the scene. However, high visual fidelity is not always required for all tasks and also adds to computational complexity. We introduce the volumetric object representation. This volumetric object representation combined with closed-set segmentation form the backbone of the Fast mode of the Bimodal 3D scene graph generation.

\subsubsection{\textbf{Volumetric object representation}}

Firstly, from each object instance segmentation mask, we extract its contour $c$, shrink it by a few pixels to mitigate noisy depth measurements near the mask boundary, and project points sampled along the contour into the world frame to obtain a 3D contour ring $C$. We denoise $C$ by removing points associated with depth jumps, where $p_i$ is removed if $\lVert p_i-p_{i-1}\rVert>d_{\max}$ and $\lVert p_{i+1}-p_i\rVert>d_{\max}$. We then reject outliers based on their distance from the median center using an $L_1$-median approximation and smooth the contour using a circular moving average. The horizontal vertices of $C$ are projected onto the $x$-$y$ plane and converted into a polygon $P$ using either a convex hull or, optionally, an $\alpha$-shape. The vertical extent is represented by ${z_{\min},z_{\max}}={\min(C_z),\max(C_z)}$. Together, $P$ and ${z_{\min},z_{\max}}$ define a polygonal prism $O$, with $P$ forming the top and bottom faces and its polygon edges defining the side faces. In practice, only $P$ and ${z_{\min},z_{\max}}$ are retained for computation; the prismatic mesh is triangulated only for visualization shown in Fig. \ref{fig:volumetricrepre}.

We adopt this prismatic representation because typical indoor objects can be approximated by compact, upright geometries, making their coarse volumetric extent useful for spatial reasoning. The representation directly supports relations such as containment and support, e.g., identifying objects within or on a table. It also facilitates merging and de-duplication across sparse views: unlike point-cloud representations such as ConceptGraphs, which may have limited point-wise correspondence across viewpoints, the contour-derived volume provides a consistent coarse geometric extent when sufficient planar structure is observed. This makes volumetric overlap a practical basis for associating observations of the same object, particularly for fast-moving robots operating with sparse views.

The same volumetric representation could be constructed from a full point cloud; however, the contour-based approach provides a computational advantage. For an $n\times m$ image mask, full mask backprojection requires $\mathcal{O}(nm)$ operations, whereas the contour contains a number of samples proportional to the image-plane perimeter. Under the assumption of compact object silhouettes, this scales as $\mathcal{O}(n+m)$. Consequently, contour-ring denoising also operates in $\mathcal{O}(n+m)$, compared with $\mathcal{O}(l^3)$ for individual point-cloud denoising of a voxelized point cloud within a cubic bounding box of side length $l$. For merging and deduplication, pairwise intersection of the simplified $x$-$y$ polygons requires $\mathcal{O}(k_i+k_j)$ operations for objects $i$ and $j$, where $k_i$ and $k_j$ are their numbers of boundary vertices. This assumes spatial-distance filtering and sparsification produce polygons with bounded local boundary complexity, such that each edge interacts with at most a constant number of edges of the opposing polygon. Under this assumption, the total number of edge interactions grows linearly with the combined boundary size.

The limitations of this volumetric approach especially in real-world is depth noise and mismatch between segmentation mask and depth image since only a sparse number of points are used. Mitigating this requires good denoising such as by averaging dense masked depth values, among others.

\subsubsection{\textbf{Merging and Deduplication}}

Objects across input frames are merged or deduplicated using their $x$-$y$ polygons and minimum/maximum $z$-coordinates. Candidate pairs are first filtered by radial $x$-$y$ distance, after which their polygon and height IoUs are evaluated. Depending on the degree of overlap and the object-class relationship, the new object may be discarded, the two objects may be averaged, or their polygons may be replaced by their union. Objects with unrelated class tags can generally be excluded from merging. The procedure is applied recursively using a tree-based representation of existing objects, allowing a newly observed object to be compared against increasingly appropriate spatial candidates without exhaustively testing all existing objects. This also accommodates cases with high $x$-$y$ overlap but little or no vertical overlap, such as identical objects positioned on vertically adjacent shelves.

\subsubsection{\textbf{Vicinity Based Object Relation}}

We construct a hierarchical organization of the object graph to provide compact local spatial context for human- or LLM-intuitive downstream landmark-based planning, which is not covered in this work. Let $O_i$ denote a 3D object and $N_i$ its corresponding node in the scene graph $\mathcal{G}_S$. For $n$ objects, we construct a symmetric visibility matrix $V_M\in\mathbb{R}^{n\times n}$, where $V_{M_{ij}}$ records the number of input frames in which $O_i$ and $O_j$ are observed together. Candidate object pairs are then filtered by a girth-buffered proximity measure, which approximates surface-to-surface proximity without computing explicit polygon-to-polygon distances. Let $d_{ij} = \left\|\mathbf{c}_i - \mathbf{c}_j\right\|_2$ denote the 3D Euclidean distance between object centroids, and let $r_h^i$ and $r_v^i$ denote the horizontal and vertical radii of $O_i$, respectively. We define the effective distance as
\begin{equation}
d_v(i,j)=
\max\left(
0,\,
d_{ij}-\max\left(
r_h^i+r_h^j,\;
r_v^i+r_v^j
\right)
\right).
\end{equation}

This provides a computationally inexpensive approximation of surface proximity: the larger combined object extent provides a conservative buffer around the centroid distance, which is appropriate for the compact, regular geometries of typical indoor furniture. Objects satisfying $d_v(i,j)<\epsilon_1$ are considered spatially proximate.

We order objects by their prismatic girth,
\begin{equation}
G_i=\sqrt[3]{\operatorname{Vol}(O_i)},
\end{equation}
and use larger objects as spatial anchors. Starting from the largest unassigned object, we recursively traverse the spatially filtered graph and assign reachable, unclaimed objects to the current vicinity while the accumulated traversal distance remains below $\epsilon_2$. This produces a partition $\mathcal{P}={P_1,\ldots,P_K}$, with each partition $P_k$ defining a vicinity $V_k$ and its initiating object serving as the central node. The hierarchy is therefore determined directly by object scale and local spatial connectivity, without introducing an additional semantic-importance metric.

The thresholds $\epsilon_1$, $\epsilon_2$, and $\epsilon_3$ respectively control local object proximity, maximum vicinity extent, and inter-vicinity connectivity. They are fixed geometric parameters rather than learned quantities and are kept constant across environments. We use 1.5, 2.5, and 3.5 m for the value respectively.  We subsequently construct a quotient visibility graph $\mathcal{G}_V$ by treating each vicinity as a supernode and aggregating visibility between constituent objects:
\begin{equation}
\tilde{w}(V_i,V_j)
=
\max_{p\in V_i,\;q\in V_j} V_{M_{pq}}.
\end{equation}
Here, $p$ and $q$ index objects belonging to vicinities $V_i$ and $V_j$, respectively. Edges may be pruned when the inter-vicinity distance exceeds $\epsilon_3$. The same procedure could be applied recursively to connected vicinities to form higher-level zones, approximately corresponding to rooms or functional areas.

\subsection{\textbf{Hybrid Representation}}

To extend the closed-set scene graph to open-set object retrieval, we augment each object node with two precomputed CLIP embeddings. The object embedding $E_O$ is obtained from the object crop, while the scene embedding $E_S$ is obtained from an expanded crop around the object to incorporate local contextual cues. The crop is empirically expanded by $20\%$ of the image width, providing nearby supporting context while limiting contamination from unrelated regions. Given a task $T$, its textual description is encoded using the CLIP text encoder to obtain the task embedding $E_T$.

Open-set retrieval compares $E_T$ with the object embeddings $E_O$ using cosine similarity. If no object-level match exceeds the current threshold $\theta_t$, the scene embeddings $E_S$ are evaluated as a contextual fallback. Since both embeddings are precomputed, this matching requires no additional LLM or VLM inference and can be performed continuously as the scene graph is updated. Thus, the Hybrid Representation provides lightweight open-set semantic evidence while retaining the closed-set scene graph representation for downstream reasoning and mode switching.

\subsection{\textbf{Mode Switching}}

While the Hybrid Representation provides inexpensive semantic evidence for direct retrieval, some tasks remain ambiguous because a CLIP similarity score alone does not establish whether an observation is relevant to the task. We therefore formulate mode-switching open-set retrieval as a sequential information-acquisition process: inexpensive evidence is gathered continuously from the scene graph, and more costly visual investigation is performed only when its expected task value warrants it. The overall procedure is summarized conceptually in Algorithm~\ref{alg}.

\begin{algorithm}[t]
\caption{Conceptual Switching-Mode Planner}
\begin{algorithmic}[1]
\Require Task $T$, scene graph stream $\mathcal{G}_S$
\Ensure Confirmed relevant objects $O{\mathrm{confirmed}}$

\State $L \gets \mathrm{LLM.decompose}(T)$
\State $E_T \gets \mathrm{CLIP.encode}(L)$
\State $O_{\mathrm{confirmed}} \gets \emptyset$

\While{$T$ not completed}
\State Receive updated scene graph $\mathcal{G}_S$

\State Compute task-conditioned object evidence $s_O$
\State $\mathcal{C} \gets$ candidates exceeding current threshold

\If{$\mathcal{C}=\emptyset$}
    \State Compute contextual scene evidence $s_S$
    \State $\mathcal{C} \gets$ candidates exceeding current threshold
\EndIf

\For{each candidate $o\in\mathcal{C}$}
    \State $a_o \gets \mathrm{LLM.decide}(o,T,\mathcal{G}_S)$
    \If{$a_o=\mathrm{inspect}$}
        \State Navigate to $o$ and obtain local VLM description $D$
        \State $o^* \gets \mathrm{LLM.confirm}(D,T)$
        \If{$o^*\neq\emptyset$}
            \State $O_{\mathrm{confirmed}}
            \gets O_{\mathrm{confirmed}}\cup\{o^*\}$
        \EndIf
    \EndIf
\EndFor

\State Update semantic evidence and exploration policy

\EndWhile

\State \Return $O_{\mathrm{confirmed}}$
\end{algorithmic}
\label{alg}
\end{algorithm}

\textbf{Phase 1 -- Task Decomposition:} Given a task $T$, the LLM decomposes the instruction into target labels $L$, which may denote specific objects or broader categories. These labels are encoded by CLIP to obtain the task embedding $E_T$. This decomposition defines the semantic hypothesis against which subsequent observations are evaluated.

\textbf{Phase 2 -- Evidence Acquisition:} For each scene-graph object $o$, we obtain two complementary task-conditioned evidence scores:
\begin{equation}
s_O(o) = \operatorname{sim}(E_T, E_O^o),
\qquad
s_S(o) = \operatorname{sim}(E_T, E_S^o).
\end{equation}
Object-level evidence is evaluated first, with scene-level evidence serving as a contextual fallback when no object-level candidate exceeds the current threshold $\theta_t$. These similarities are treated as relative evidence scores rather than calibrated probabilities. Candidates exceeding $\theta_t$ are passed to the planner as hypotheses for further action.

\textbf{Phase 3 -- Information-Acquisition Decision:} A candidate is not treated as a confirmed detection. Instead, the LLM planner determines whether resolving the remaining uncertainty is warranted given the task, possible costs incurred, scene context, and accumulated evidence. This admits a decision-theoretic, value-of-information interpretation:
\begin{equation}
\operatorname{Inspect}(o)
\iff
\operatorname{VOI}(o \mid T, \mathcal{G}_S)
>
C_{\mathrm{nav}}
+
C_{\mathrm{VLM}}
+
C_{\mathrm{seg}},
\end{equation}
where $C_{\mathrm{nav}}$, $C_{\mathrm{VLM}}$, and $C_{\mathrm{seg}}$ denote navigation, VLM inference, and open-set segmentation costs, respectively. In our implementation, navigation requires approximately $3$--$5$ s and VLM inference approximately $3$ s, while segmentation is incurred only when required. The formulation is an interpretation of the decision process rather than an explicit numerical VOI calculation; the LLM approximates the corresponding task-dependent policy using the candidate evidence, scene context, and task requirements.

When inspection is selected, the robot navigates to $o$ and obtains a local VLM description $D$. The LLM then evaluates the observation against the task:
\begin{equation}
o^* = \operatorname{LLM.confirm}(D,T),
\end{equation}
Confirmed objects are added to the relevant-object set and, when required, segmented using the open-set model and incorporated into the scene graph.

The candidate threshold is adapted throughout exploration to balance inspection effort against the risk of overlooking weakly expressed targets. Repeated rejection of CLIP-generated candidates indicates that the threshold is too permissive and therefore increases it,
\begin{equation} \theta_{t+1}
=
\min\left( \theta_{\max}, \theta_t + \Delta_+ \right), \end{equation}
whereas persistent absence of useful candidates suggests that it is too restrictive and gradually lowers it,
\begin{equation}
\theta_{t+1}
=
\max\left(
\theta_{\min},
\theta_t - \Delta_-
\right),
\qquad
\Delta_- < \Delta_+.
\end{equation}
Relaxation is gated by a counter, while previously observed evidence is temporally decayed to prevent stale hypotheses from dominating newly acquired candidates.

Thus, BiMoSG uses CLIP as low-cost continuous evidence and selectively invokes LLM--VLM inspection when additional information is warranted, allowing computation and robot motion to be allocated according to task requirements. We use GPT-4o \cite{hurst2024gpt} as the LLM planner and SmolVLM2 \cite{marafioti2025smolvlm} for local visual description.

The three BiMoSG modes progressively support more complex tasks (Table~\ref{tab:capability_matrix}): Scene Graph Representation (SGR) handles closed-set retrieval, Hybrid Mode supports open-set retrieval, and Mode Switching enables complex open-set tasks through LLM--VLM reasoning. Thus, BiMoSG progresses from semantic matching to visual reasoning.
\begin{table}[t]
\centering
\caption{Capability matrix of BiMoSG.}
\label{tab:capability_matrix}
\scriptsize
\setlength{\tabcolsep}{3pt}
\renewcommand{\arraystretch}{0.75}
\begin{tabular}{lccc}
\hline
\textbf{Task} & \textbf{SGR} & \textbf{Hybrid} & \textbf{Mode Switch} \\
\hline
Closed-set  & \checkmark & \checkmark & \checkmark \\
Open-set    & --         & \checkmark & \checkmark \\
Complex Open-set  & --         & --         & \checkmark \\
\hline
\end{tabular}
\end{table}

ded to the confirmed set; otherwise, the planner continues monitoring the scene graph.

\section{Experiments and Results}

\subsection{\textbf{Hardware Detail}}
All experiments using the Replica dataset and simulator were conducted on a computer equipped with an NVIDIA RTX 4090 GPU, on a 24-core Intel i9-13900 CPU, while the real-world experiments were conducted using an NVIDIA Jetson AGX Thor.

\vspace{-0.005\textheight}

\subsection{\textbf{Scene Graph Geometrical Representation}}
We first evaluate the performance of the scene graph generation component independently of the task. This evaluation focuses on two aspects of the proposed prismatic volumetric representation: (a) \textbf{Geometric and semantic consistency}  (b) \textbf{Computational efficiency}. Object retrieval is excluded from this evaluation to isolate the performance of the scene graph generation component.

\subsubsection{\textbf{Geometric and Semantic Consistency Comparison}}

\begin{figure}[H]
\centering
\includegraphics[width=0.8\columnwidth]{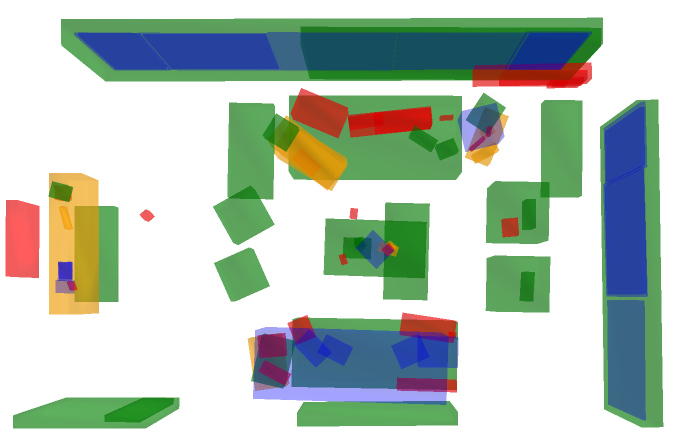}
\captionof{figure}{Geometrical projection accuracy visualization for Replica dataset(room0). Topdown view - left: table; middle: sofa and coffee table; right: sofas and windows. Green - match bounding box and class; Orange - match bounding box; Red : False positive; Blue : False negative}
\label{fig:geometrical acc}
\end{figure}

Our scene graph employs a prismatic volumetric representation, which makes direct geometric benchmarking more challenging than representations based on point clouds or detailed meshes. To evaluate the geometric and semantic consistency of the proposed representation, we perform volumetric matching against the semantic ground truth of the Replica Dataset~\cite{straub2019replicadatasetdigitalreplica}. Specifically, we compare the oriented bounding boxes (OBBs) of our predicted volumes with the corresponding OBBs derived from the Replica ground truth. Table~\ref{table:repproj} summarizes the geometrical projection accuracy across eight Replica environments, while Fig. \ref{fig:geometrical acc} illustrates the corresponding results visualization for the Room 0. We do not compare with baselines as baselines measure point cloud segmentation accuracy rather than volumetric.

The relatively lower precision is primarily attributed to small segmentation artifacts and occasional incorrect class predictions from the closed-set segmentation module. Recall measures the proportion of relevant ground-truth objects successfully matched by the scene graph, while linearized mIoU uses the cube root of volumetric agreement between predicted and ground-truth OBBs, reducing the sensitivity of box volume to small geometric deviations. Semantic accuracy measures the correctness of predicted classes among geometrically matched objects, thereby separating semantic errors from geometric matching. Some discrepancies also arise from mismatches between the ground truth and our representation; for example, the window at the left of Fig.~4 is detected but absent from the ground truth, while the lamp OBB is overestimated in the ground truth by more than $10\times$. Overall, the results demonstrate a consistent geometric and semantic representation.
\begin{table}[H]
\centering
\caption{Geometric and semantic consistency metrics for BiMoSG across several Replica environments.}
\resizebox{\columnwidth}{!}{
\begin{tabular}{l|ccccc}
\hline
\textbf{Replica Dataset} & \textbf{Precision} & \textbf{Recall} & \textbf{Semantic Acc} & \textbf{Linearized mIoU}  \\
\hline

Room 0   & 0.594 & 0.613 & 0.763 & 0.627 \\
Room 1   & 0.449 & 0.595 & 0.636 & 0.606  \\
Room 2   & 0.424 & 0.582 & 0.600 & 0.639  \\
Office 0 & 0.388 & 0.759 & 0.316 & 0.604  \\
Office 1 & 0.429 & 0.857 & 0.444 & 0.607  \\
Office 2 & 0.579 & 0.690 & 0.552 & 0.591  \\
Office 3 & 0.483 & 0.558 & 0.759 & 0.619  \\
Office 4 & 0.373 & 0.633 & 0.667 & 0.619  \\
\hline
\textbf{Average}  & \textbf{0.466} & \textbf{0.661} & \textbf{0.592} & \textbf{0.610} \\
\hline
\label{table:repproj}

\end{tabular}
}
\end{table}

\vspace{-0.03\textheight}

\subsubsection{\textbf{Computational Efficiency}}

We next evaluate the computational efficiency of the proposed scene graph representation by comparing its processing time with existing scene graph generation approaches on the Replica Dataset~\cite{straub2019replicadatasetdigitalreplica}. We compare against two representative methods, ConceptGraphs and BBQ, using the same evaluation setting. To isolate the computational cost of scene graph generation, we measure only the operations required to construct and update the geometric scene representation, including backprojection of segmentation mask, point cloud, mesh or volume formation, temporal merging and deduplication, and related geometric processing. We exclude any open-set element (segmentation, CLIP, VLM, LLM, open-set tasks) and any time not spent on geometric representation, from this computation. Thus, we evaluate the geometric representation for fair benchmarking. We do not compare retrieval accuracy or task success, as ConceptGraphs and BBQ are not designed for online, real-time scene graph generation.

As summarized in Table \ref{tab:running_time_camparisons}, our volumetric representation substantially reduces the processing time compared with the point cloud and mesh based baselines. On average, our approach achieves about a two fold reduction in time compared with BBQ and a four-fold reduction compared with ConceptGraphs. Again, this does not indicate the speed of task completion or overall open-set scene graph construction.

\begin{table}[!htb]
\centering

\caption{Scene graph generation time  comparison with baselines}

{\scriptsize{
\begin{tabular}{lccc}

\hline
\textbf{Replica Dataset} & \textbf{ConceptGraphs (sec)} & \textbf{BBQ (sec)} & \textbf{Our Method (sec)}\\
\hline

Room0   & 133.0460  & 56.5553  & 33.59 \\
Room1   & 88.7586  & 45.9022 & 21.28 \\
Room2   & 68.1647  & 53.0010 & 27.34 \\
office0   & 48.9914 & 35.6428 & 13.68 \\
office1   & 34.9416  & 29.7156 & 14.77 \\
office2   & 81.4436 & 43.0160 & 20.64 \\
office3  & 89.7984  & 52.3345 & 26.56 \\
office4   & 88.6095   & 48.1167 & 17.20 \\
\hline
\textbf{Avg per frame}   & \textbf{0.4003}   & \textbf{0.2277} & \textbf{0.1094} \\

\hline
\end{tabular}}}
\label{tab:running_time_camparisons}
\end{table}

\vspace{-0.02\textheight}
\subsection{\textbf{Bimodal Scene Graph Generation}}

\subsubsection{\textbf{Experimental Setup}}
We evaluated BiMoSG on two categories of tasks, corresponding to different levels of abstraction. These include: (a) \textbf{Simple Object Retrieval:} The agent is asked to locate a specific object instance in the environment. \emph{Examples:} ``Find a book,'' ``Find a bicycle.''; (b) \textbf{Complex Object Retrieval:} The agent is asked to interpret a functional or contextual request, which requires reasoning about object affordances or placement rather than simply retrieving a named item. Ex., ``I need to repair a lab instrument'', whose target objects can vary vastly, but categories can be deduced: ``books, manuals'', ``workshop tools'', and ``lab instruments''. The exact objects retrieved, e.g. ``microscope repair manual'', ``allen keys'', and ``microscope'' depend on the objects available in the scene rather than guessed in advance.

\subsubsection{\textbf{Evaluation Metrics}}
To evaluate the performance of BiMoSG, we consider the following metrics: (a) \textbf{Time to Completion:} Total time taken to execute a high-level task, measuring efficiency of planning and perception and (b) \textbf{False Positive Rate:} Number of false positives encountered before completing a task. As task evaluation depends on factors difficult to factor and variation between approaches, we use false positive rate as a proxy for semantic accuracy (related to a task). However, we do not directly compare 2D open-set segmentation labeling as it is a constituent part and not the focus of our work.

\begin{table*}[t]

\centering
\caption{Performance comparison across Small, Medium, and Large habitat environments.}
\label{tab:habitat-combined}
\setlength{\tabcolsep}{4pt}

\resizebox{\textwidth}{!}{
\begin{tabular}{c|l ccc ccc|l ccc ccc|l ccc ccc}
\hline
& \multicolumn{7}{c|}{\textbf{Small}} & \multicolumn{7}{c|}{\textbf{Medium}} & \multicolumn{7}{c}{\textbf{Large}} \\
\cline{2-22}
\textbf{Task} & \textbf{Class} & \multicolumn{3}{c}{\textbf{Clio}} & \multicolumn{3}{c|}{\textbf{Ours}}
& \textbf{Class} & \multicolumn{3}{c}{\textbf{Clio}} & \multicolumn{3}{c|}{\textbf{Ours}}
& \textbf{Class} & \multicolumn{3}{c}{\textbf{Clio}} & \multicolumn{3}{c}{\textbf{Ours}} \\
& & \textbf{Time (mm:ss)} & \textbf{FP} & \textbf{Succ.} & \textbf{Time (mm:ss)} & \textbf{FP} & \textbf{Succ.}
& & \textbf{Time (mm:ss)} & \textbf{FP} & \textbf{Succ.} & \textbf{Time (mm:ss)} & \textbf{FP} & \textbf{Succ.}
& & \textbf{Time (mm:ss)} & \textbf{FP} & \textbf{Succ.} & \textbf{Time (mm:ss)} & \textbf{FP} & \textbf{Succ.} \\
\hline

Open
& Keyboard        & 1:46 & 22 & \textbf{Y} & 2:08 & \textbf{1} & N
& Ceiling Fan     & 2:34 & 15 & Y & \textbf{0:49} & \textbf{0} & Y
& Broom           & 1:17 & 12 & \textbf{Y} & \textbf{0:40} & \textbf{3} & N \\

Open
& Ceiling Fan     & 1:11 & 11 & Y & \textbf{0:50} & \textbf{3} & Y
& Printer         & 1:02 & \textbf{0} & N & \textbf{0:30} & 2 & \textbf{Y}
& Bathtub         & 2:02 & 11 & Y & \textbf{1:28} & \textbf{3} & Y \\

Open
& Bathtub         & 1:49 & 16 & Y & \textbf{1:40} & \textbf{1} & Y
& Shoes           & 4:10 & 24 & Y & \textbf{1:24} & \textbf{3} & Y
& Chess painting  & 2:30 & 11 & Y & \textbf{1:09} & \textbf{3} & Y \\

Open
& Kitchen exhaust & 1:10 & 19 & Y & \textbf{0:52} & \textbf{2} & Y
& Coffee machine  & 2:14 & 13 & Y & \textbf{1:12} & \textbf{0} & Y
& Beach hat       & 1:41 & 12 & Y & \textbf{0:40} & \textbf{0} & Y \\

Open
& Toilet Paper    & 0:44 & 21 & Y & 0:44 & \textbf{2} & Y
& Christmas tree  & 3:02 & 8  & Y & \textbf{2:21} & \textbf{0} & Y
& Unicorn head    & 2:11 & 16 & Y & \textbf{0:50} & \textbf{1} & Y \\

\hline

Close
& Ottoman         & 1:13 & 13 & Y & \textbf{0:25} & \textbf{0} & Y
& Painting        & 1:48 & 17 & Y & \textbf{1:25} & \textbf{0} & Y
& Monitor         & 1:49 & 12 & Y & \textbf{0:55} & \textbf{2} & Y \\

Close
& Mirror          & 1:19 & 23 & Y & \textbf{1:18} & \textbf{3} & Y
& Bathtub         & 2:25 & 32 & Y & \textbf{1:42} & \textbf{3} & Y
& Mirror          & 1:23 & 11 & Y & \textbf{0:40} & \textbf{2} & Y \\

Close
& Sofa            & 0:56 & 11 & Y & \textbf{0:50} & \textbf{3} & Y
& Sofa            & 0:41 & 7  & Y & \textbf{0:40} & \textbf{0} & Y
& Washing machine & 3:12 & 2  & Y & \textbf{1:17} & \textbf{0} & Y \\

Close
& Bed             & 1:17 & 24 & Y & \textbf{1:15} & \textbf{3} & Y
& Bed             & 2:51 & 22 & Y & \textbf{1:49} & \textbf{3} & Y
& Dinning table   & 1:18 & 3  & Y & \textbf{1:05} & \textbf{1} & Y \\

Close
& Refrigerator    & 1:04 & 16 & Y & \textbf{0:30} & \textbf{1} & Y
& Refrigerator    & 2:56 & 14 & Y & \textbf{0:54} & \textbf{2} & Y
& Bed             & 3:14 & 18 & Y & \textbf{1:25} & \textbf{6} & Y \\

\hline
\end{tabular}
}
\end{table*}

\subsubsection{\textbf{Simple Object Retrieval}}

\textit{A note on benchmark comparison.} Although ASHiTA addresses a closely related task-driven hierarchical scene representation problem, its SG3D evaluation assumes predefined task hierarchies over reconstructed scenes, whereas our evaluation focuses on online discovery and organization of an initially unknown scene. We therefore omit a direct quantitative comparison due to the fundamentally different evaluation settings.

We simulate and run experiments in several environments to evaluate BiMoSG (Hybrid Representation) and Clio performing tasks on the Habitat Matterport3D dataset \cite{chang2017matterport3d}. We used Clio-online as recommended by the authors as it is a real-time execution task. The results are summarized in Table \ref{tab:habitat-combined}. We selected three scenes of varying sizes: small (00801-HaxA7YrQdEC) with 9 rooms, medium (00802-wcojb4TFT35) with 15 rooms, and large (00809-Qpor2mEya8F) with 21 rooms. The scenes primarily correspond to hallway, living room, kitchen, bedroom, bathroom and garage areas. We run various open-set and closed-set class tasks and report the time taken and false positive rate in these tasks. The simulation is controlled by a human controller, at a slightly brisk pace, and has the same starting point, end point and trajectory for both setups.

Referring to Table\ref{tab:habitat-combined}, where $FP$ denotes false positive, We observe that our approach generally achieves lower execution times and fewer false positives across the evaluated tasks. We have used the default settings provided in the repository. We have stuck to these defaults as tweaking the parameters did not significantly change the results. 

We attribute the difference in execution time primarily to object registration. BiMoSG registers objects immediately upon observation, whereas Clio-online aggregates registration over several samples (3--5) to improve consistency. Thus, despite following the same trajectories, Clio-online spends more time accumulating observations necessary for object registration. This reflects a deliberate trade-off between registration consistency and the faster, coarse geometric representation used by BiMoSG. The difference in false-positive rate arises primarily from the differing objectives of the two systems. BiMoSG hybrid mode performs task-specific object confirmation using an absolute CLIP similarity threshold, whereas Clio is designed to organize and compress task-relevant objects and places, rather than directly confirm a fixed set of target objects. Clio-online additionally operates under incremental uncertainty, whereas batch processing can exploit observations accumulated over the complete scene. The online setting may therefore admit more transient false positives during exploration, including those arising from reconstruction issues. In contrast, BiMoSG uses a empirically determined similarity threshold ($0.26$ in our experiments), which provides effective filtering under typical conditions but can be less robust to degraded visual conditions such as poor visibility, unusual lighting.

\subsubsection{\textbf{Complex Object Retrieval}}

To evaluate BiMoSG (Mode Switching) on complex object retrieval, we conducted several high-level object retrieval experiments; we highlight the \textit{Vegetables Grocery} task as a representative example, shown in Fig.~\ref{fig:open-example}. Given the instruction, ``I just bought vegetables and milk, what should I do?'', the planner decomposes the task into labels corresponding to likely storage locations (Phase 1), identifies several candidate objects via CLIP similarity matching (Phase 2), and confirms each candidate's relevance following VLM description and LLM verification, with the countertop confirmed. (Phase 3). Since existing scene graph methods do not incorporate a comparable mode-switching mechanism, a direct baseline comparison for this task is not available. Execution time for this and other simple and complex retrieval experiments is summarized in Table~\ref{tab:real_results_01}.

\begin{table}[t]
\centering
\caption{Mode Switching experiments}
\label{tab:real_results_01}
\scriptsize
\setlength{\tabcolsep}{2pt}
\renewcommand{\arraystretch}{0.8}

\begin{tabular}{l l c}  
\hline
Task Type & Task & Execution Time (s) \\
\hline

\multirow{2}{*}{Simple} 
    & Trash Can       & 57 \\ 
    & Kitchen Island  & 23 \\ 
\hline

\multirow{2}{*}{Complex} 
    & Ice Cream           & 19 \\ 
    & Vegetables Grocery  & 45 \\ 
\hline
\end{tabular}
\end{table}

\subsubsection{\textbf{Real-World Experiments}}

\begin{figure}[t]
\centering

\includegraphics[width=\columnwidth]{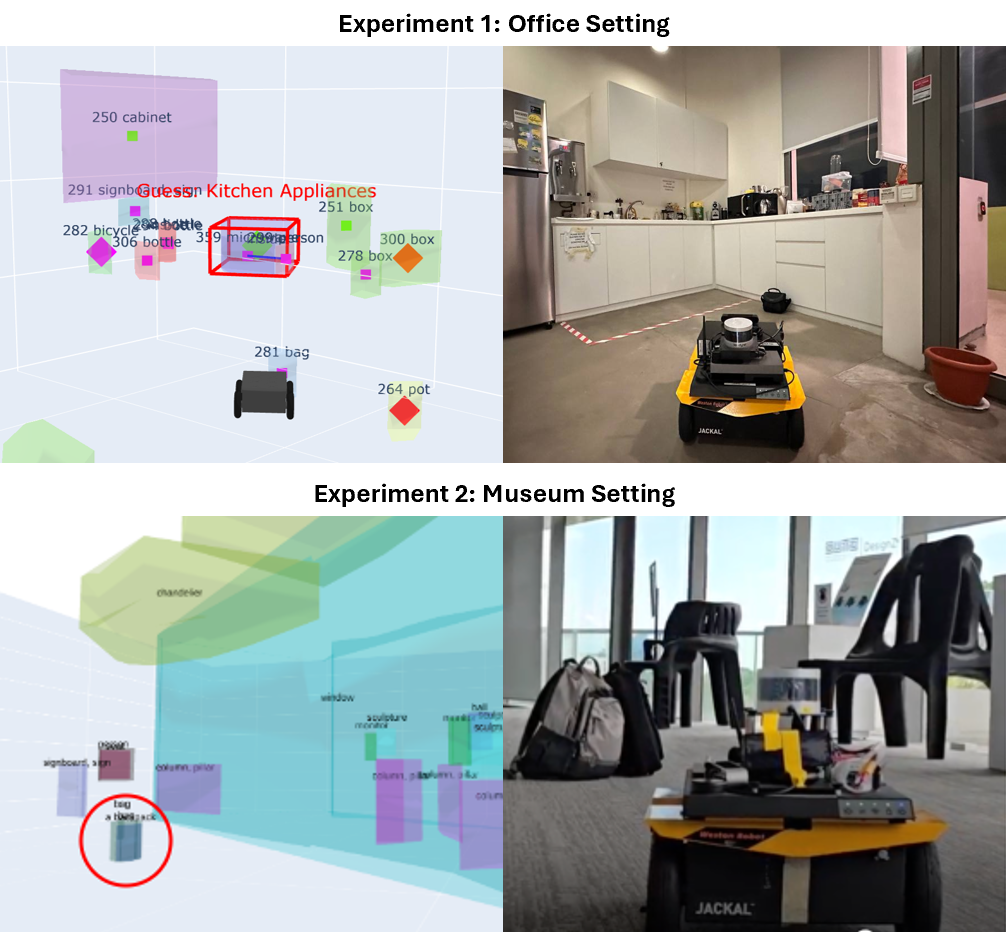}

\caption{Real-world experiments on a Clearpath Jackal robot for complex object retrieval. \textbf{Top:} food-heating task in a lab pantry, where the target object is identified as a microwave. \textbf{Bottom:} security-inspection task in a museum , where the target object is identified as a backpack.}
\label{fig:real_results}
\end{figure}

To evaluate BiMoSG's (Mode switching) capability on complex object retrieval tasks in the real world, we deploy it on a Clearpath Jackal robot across two environments: a museum and an office. 
In the museum environment, the robot is given the instruction, ``You are a security robot. Look for things that seem out of place and suspicious.'' The planner decomposes the task into the target label \textit{suspicious objects} (Phase 1), and identifies a candidate object in the scene graph via CLIP similarity matching (Phase 2). The VLM then describes the local scene at the candidate location, and the LLM confirms a bag as the relevant object, adding it to the confirmed set (Phase 3). In the pantry environment, the robot is given the instruction, ``I am hungry, how do I heat up my food?'' The planner decomposes the task into the target label \textit{kitchen appliances} (Phase 1), identifies a candidate match via CLIP similarity (Phase 2), and confirms the object as a microwave following VLM description and LLM verification (Phase 3). Fig.~\ref{fig:real_results} shows both real-world experiments along with the corresponding scene graphs generated by BiMoSG.

\section{Conclusion}

We presented BiMoSG, a bimodal 3D scene graph framework that switches between a lightweight  \textit{Fast Mode} and a slower, LLM--VLM-driven \textit{Mode Switching} mechanism, escalating only when task-related reasoning is required. In simulation, BiMoSG's fast mode generates scene graphs substantially faster than existing open-vocabulary baselines, while its mode-switching mechanism achieves lower execution time and fewer false positives than a state-of-the-art baseline on object retrieval tasks. We further validated BiMoSG through real-world deployment on a mobile robot in complex retrieval tasks in unknown environments, demonstrating its viability as a real-time, on-the-edge scene graph generator for robot perception and task-driven reasoning.

\bibliographystyle{IEEEtran}

\bibliography{root_main}

@inproceedings{gu2024conceptgraphs,
  title={Conceptgraphs: Open-vocabulary 3d scene graphs for perception and planning},
  author={Gu, Qiao and Kuwajerwala, Ali and Morin, Sacha and Jatavallabhula, Krishna Murthy and Sen, Bipasha and Agarwal, Aditya and Rivera, Corban and Paul, William and Ellis, Kirsty and Chellappa, Rama and others},
  booktitle={2024 IEEE International Conference on Robotics and Automation (ICRA)},
  pages={5021--5028},
  year={2024},
  organization={IEEE}
}

@misc{gorlo2025moment,
      title={Describe Anything Anywhere At Any Moment}, 
      author={Nicolas Gorlo and Lukas Schmid and Luca Carlone},
      year={2025},
      eprint={2512.00565},
      archivePrefix={arXiv},
      primaryClass={cs.CV},
      url={https://arxiv.org/abs/2512.00565}, 
}

@misc{cheng2022maskedattentionmasktransformeruniversal,
      title={Masked-attention Mask Transformer for Universal Image Segmentation}, 
      author={Bowen Cheng and Ishan Misra and Alexander G. Schwing and Alexander Kirillov and Rohit Girdhar},
      year={2022},
      eprint={2112.01527},
      archivePrefix={arXiv},
      primaryClass={cs.CV},
      url={https://arxiv.org/abs/2112.01527}, 
}

@book{kahneman2011thinking,
  address = {New York},
  author = {Kahneman, Daniel},
  description = {Thinking, Fast and Slow: Amazon.de: Daniel Kahneman: Englische Bücher},
  isbn = {9780374275631 0374275637},
  publisher = {Farrar, Straus and Giroux},
  refid = {706020998},
  title = {Thinking, fast and slow},
  url = {https://www.amazon.de/Thinking-Fast-Slow-Daniel-Kahneman/dp/0374275637/ref=wl_it_dp_o_pdT1_nS_nC?ie=UTF8&colid=151193SNGKJT9&coliid=I3OCESLZCVDFL7},
  year = 2011
}

@article{maggio2024clio,
  title={Clio: Real-time task-driven open-set 3d scene graphs},
  author={Maggio, Dominic and Chang, Yun and Hughes, Nathan and Trang, Matthew and Griffith, Dan and Dougherty, Carlyn and Cristofalo, Eric and Schmid, Lukas and Carlone, Luca},
  journal={IEEE Robotics and Automation Letters},
  year={2024},
  publisher={IEEE}
}

@article{jatavallabhula2023conceptfusion,
  title={Conceptfusion: Open-set multimodal 3d mapping},
  author={Jatavallabhula, Krishna Murthy and Kuwajerwala, Alihusein and Gu, Qiao and Omama, Mohd and Chen, Tao and Maalouf, Alaa and Li, Shuang and Iyer, Ganesh and Saryazdi, Soroush and Keetha, Nikhil and others},
  journal={arXiv preprint arXiv:2302.07241},
  year={2023}
}

@INPROCEEDINGS{Werby-RSS-24, 
      AUTHOR    = {Abdelrhman Werby AND Chenguang Huang AND Martin Büchner AND Abhinav Valada AND Wolfram Burgard}, 
      TITLE     = {{Hierarchical Open-Vocabulary 3D Scene Graphs for Language-Grounded Robot Navigation}}, 
      BOOKTITLE = {Proceedings of Robotics: Science and Systems}, 
      YEAR      = {2024}, 
      ADDRESS   = {Delft, Netherlands}, 
      MONTH     = {July}, 
      DOI       = {10.15607/RSS.2024.XX.077} 
}

@inproceedings{kirillov2023segment,
  title={Segment anything},
  author={Kirillov, Alexander and Mintun, Eric and Ravi, Nikhila and Mao, Hanzi and Rolland, Chloe and Gustafson, Laura and Xiao, Tete and Whitehead, Spencer and Berg, Alexander C and Lo, Wan-Yen and others},
  booktitle={Proceedings of the IEEE/CVF international conference on computer vision},
  pages={4015--4026},
  year={2023}
}

@inproceedings{liu2023llava,
    author      = {Liu, Haotian and Li, Chunyuan and Wu, Qingyang and Lee, Yong Jae},
    title       = {Visual Instruction Tuning},
    booktitle   = {NeurIPS},
    year        = {2023}
}

@inproceedings{radford2021learning,
  title={Learning transferable visual models from natural language supervision},
  author={Radford, Alec and Kim, Jong Wook and Hallacy, Chris and Ramesh, Aditya and Goh, Gabriel and Agarwal, Sandhini and Sastry, Girish and Askell, Amanda and Mishkin, Pamela and Clark, Jack and others},
  booktitle={International conference on machine learning},
  pages={8748--8763},
  year={2021},
  organization={PMLR}
}

@inproceedings{rosinol2020kimera,
  title={Kimera: an open-source library for real-time metric-semantic localization and mapping},
  author={Rosinol, Antoni and Abate, Marcus and Chang, Yun and Carlone, Luca},
  booktitle={2020 IEEE International Conference on Robotics and Automation (ICRA)},
  pages={1689--1696},
  year={2020},
  organization={IEEE}
}

@inproceedings{hughes2022hydra,
  title={Hydra: A Real-time Spatial Perception System for 3D Scene Graph Construction and Optimization},
  author={Hughes, Nathan and Chang, Yun and Carlone, Luca},
  booktitle={Robotics: Science and Systems},
  year={2022}
}

@inproceedings{peng2023openscene,
  title={Openscene: 3d scene understanding with open vocabularies},
  author={Peng, Songyou and Genova, Kyle and Jiang, Chiyu and Tagliasacchi, Andrea and Pollefeys, Marc and Funkhouser, Thomas and others},
  booktitle={Proceedings of the IEEE/CVF conference on computer vision and pattern recognition},
  pages={815--824},
  year={2023}
}

@inproceedings{koch2024open3dsg,
  title={Open3dsg: Open-vocabulary 3d scene graphs from point clouds with queryable objects and open-set relationships},
  author={Koch, Sebastian and Vaskevicius, Narunas and Colosi, Mirco and Hermosilla, Pedro and Ropinski, Timo},
  booktitle={Proceedings of the IEEE/CVF Conference on Computer Vision and Pattern Recognition},
  pages={14183--14193},
  year={2024}
}

@article{hurst2024gpt,
  title={Gpt-4o system card},
  author={Hurst, Aaron and Lerer, Adam and Goucher, Adam P and Perelman, Adam and Ramesh, Aditya and Clark, Aidan and Ostrow, AJ and Welihinda, Akila and Hayes, Alan and Radford, Alec and others},
  journal={arXiv preprint arXiv:2410.21276},
  year={2024}
}

@misc{straub2019replicadatasetdigitalreplica,
      title={The Replica Dataset: A Digital Replica of Indoor Spaces}, 
      author={Julian Straub and Thomas Whelan and Lingni Ma and Yufan Chen and Erik Wijmans and Simon Green and Jakob J. Engel and Raul Mur-Artal and Carl Ren and Shobhit Verma and Anton Clarkson and Mingfei Yan and Brian Budge and Yajie Yan and Xiaqing Pan and June Yon and Yuyang Zou and Kimberly Leon and Nigel Carter and Jesus Briales and Tyler Gillingham and Elias Mueggler and Luis Pesqueira and Manolis Savva and Dhruv Batra and Hauke M. Strasdat and Renzo De Nardi and Michael Goesele and Steven Lovegrove and Richard Newcombe},
      year={2019},
      eprint={1906.05797},
      archivePrefix={arXiv},
      primaryClass={cs.CV},
      url={https://arxiv.org/abs/1906.05797}, 
}

@inproceedings{linok2025beyond,
  title={Beyond bare queries: Open-vocabulary object grounding with 3d scene graph},
  author={Linok, Sergey and Zemskova, Tatiana and Ladanova, Svetlana and Titkov, Roman and Yudin, Dmitry and Monastyrny, Maxim and Valenkov, Aleksei},
  booktitle={2025 IEEE International Conference on Robotics and Automation (ICRA)},
  pages={13582--13589},
  year={2025},
  organization={IEEE}
}

@inproceedings{chang2025ashita,
  title={ASHiTA: Automatic scene-grounded hierarchical task analysis},
  author={Chang, Yun and Fermoselle, Leonor and Ta, Duy and Bucher, Bernadette and Carlone, Luca and Wang, Jiuguang},
  booktitle={Proceedings of the Computer Vision and Pattern Recognition Conference},
  pages={29458--29468},
  year={2025}
}

@article{zhao2023fast,
  title={Fast segment anything},
  author={Zhao, Xu and Ding, Wenchao and An, Yongqi and Du, Yinglong and Yu, Tao and Li, Min and Tang, Ming and Wang, Jinqiao},
  journal={arXiv preprint arXiv:2306.12156},
  year={2023}
}

@article{marafioti2025smolvlm,
  title={SmolVLM: Redefining small and efficient multimodal models}, 
  author={Andrés Marafioti and Orr Zohar and Miquel Farré and Merve Noyan and Elie Bakouch and Pedro Cuenca and Cyril Zakka and Loubna Ben Allal and Anton Lozhkov and Nouamane Tazi and Vaibhav Srivastav and Joshua Lochner and Hugo Larcher and Mathieu Morlon and Lewis Tunstall and Leandro von Werra and Thomas Wolf},
  journal={arXiv preprint arXiv:2504.05299},
  year={2025}
}

@article{zhang2023faster,
  title={Faster segment anything: Towards lightweight sam for mobile applications},
  author={Zhang, Chaoning and Han, Dongshen and Qiao, Yu and Kim, Jung Uk and Bae, Sung-Ho and Lee, Seungkyu and Hong, Choong Seon},
  journal={arXiv preprint arXiv:2306.14289},
  year={2023}
}

@misc{chang2017matterport3d,
      title={Matterport3D: Learning from RGB-D Data in Indoor Environments}, 
      author={Angel Chang and Angela Dai and Thomas Funkhouser and Maciej Halber and Matthias Nießner and Manolis Savva and Shuran Song and Andy Zeng and Yinda Zhang},
      year={2017},
      eprint={1709.06158},
      archivePrefix={arXiv},
      primaryClass={cs.CV},
      url={https://arxiv.org/abs/1709.06158}, 
}

@inproceedings{carion2026sam,
  title={Sam 3: Segment anything with concepts},
  author={Carion, Nicolas and Gustafson, Laura and Hu, Yuan-Ting and Debnath, Shoubhik and Hu, Ronghang and Suris Coll-Vinent, Didac and Ryali, Chaitanya and Alwala, Kalyan Vasudev and Khedr, Haitham and Huang, Andrew and others},
  booktitle={International conference on learning representations},
  volume={2026},
  pages={138846--138923},
  year={2026}
}

\end{document}